\documentclass[a4paper]{article}

\usepackage{INTERSPEECH2015} 

\usepackage{graphicx}
\usepackage{amssymb,amsmath,bm}
\usepackage{textcomp}
\usepackage{arydshln}

 \ninept 

\title{Speaker-independent machine lip-reading with speaker-dependent viseme classifiers}

\makeatletter
\def\name#1{\gdef\@name{#1\\}}
\makeatother \name{{\em Helen L. Bear$^1$, Stephen J. Cox$^1$, Richard W. Harvey$^1$ }}

 \address{$^1$University of East Anglia, UK\\ 
  {\small \tt \{helen.bear,r.w.harvey,s.j.cox\}@uea.ac.uk} } 

\begin{document}

  \maketitle
  \begin{abstract}
  In machine lip-reading, which is identification of speech from visual-only information, there is evidence to show that visual speech is highly dependent upon the speaker \cite{cox2008challenge}. Here, we use a phoneme-clustering method to form new phoneme-to-viseme maps for both individual and multiple speakers. We use these maps to examine how similarly speakers talk visually. We conclude that broadly speaking, speakers have the same repertoire of mouth gestures, where they differ is in the use of the gestures.
\end{abstract}

  \section{Introduction}
  
Speaker identity is known to be important in the recognition of speech from visual-only information (lip-reading) \cite{cox2008challenge}, more so than in audio speech. One of the difficulties in dealing with visual speech is what the findamental units for recognition should be.  The term {\em viseme} is loosely defined \cite{fisher1968confusions} to mean a visually indistinguishable unit of speech, and a set of visemes is usually defined by grouping together a number of phonemes that have a (supposedly) indistinguishable visual appearance. Several many-to-one mappings from phonemes to visemes have been proposed and investigated \cite{chen1998audio}, \cite{fisher1968confusions} or \cite{Hazen1027972}. In \cite{bear2014phoneme}, a new idea of using speaker-dependent visemes is presented. The method can be summarised
as follows:  
\begin{enumerate}
\item Perform speaker-dependent phoneme recognition with recognisers that use phoneme units. 
\item By aligning the phoneme output of the recogniser with the transcription of the word uttered, a confusion matrix for each speaker is produced detailing which phonemes are confused with which others. 
\item Phonemes are clustered into groups (visemes) based on the confusions identified in step two. The clustering algorithm permits phonemes to be grouped into a single viseme, $V$ only if each phoneme has been confused with all the others within $V$.  Consonant and vowel phonemes are not permitted to be mixed within a viseme class. The result of this process is a Phoneme-to-Viseme (P2V) map $M$ for each speaker---for further details, see \cite{bear2014phoneme}. 
\item These new speaker-dependent viseme sets are then used as units for visual speech recognition for a speaker.
\end{enumerate}
This resulted in a small improvement in speaker-dependent recognition \cite{bear2014phoneme}. The question then arises to what extent such maps are independent of the speaker, and if so, how speaker independence might be examined. In particular, we are interested in the interaction between the data used to train the models and the viseme classes themselves.  

 \section{Dataset description}
\begin{figure*}[!pht]
\centering
\includegraphics[width=0.8\textwidth]{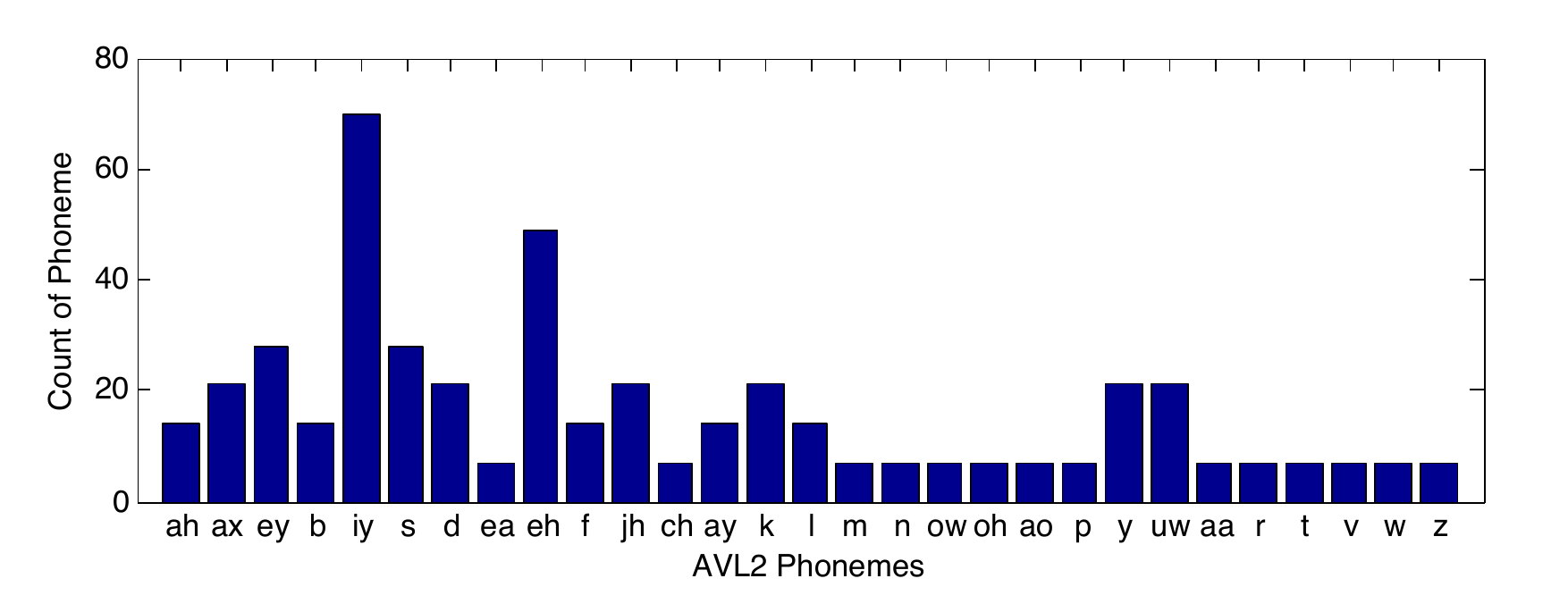}
\caption{Phoneme histogram of AVLetters-2 dataset}
\label{fig:histogram}
\end{figure*}

We use the AVLetters2 (AVL2) dataset \cite{cox2008challenge}, to train and test recognisers based upon the new P2V mappings. This dataset consists of four British-English speakers reciting the alphabet seven times. The full-faces of the speakers are tracked using Active Appearance Models (AAMs)~\cite{Matthews_Baker_2004} from which lip-only combined shape and appearance features are extracted. We select AAM features because they are known to out-perform other feature methods in machine visual-only lip-reading~\cite{cappelletta2012phoneme}. Figure~\ref{fig:histogram} shows the count of the 29 phonemes that appear in the phoneme transcription of AVL2, allowing for duplicate pronunciations, (with the silence phoneme omitted). The BEEP pronunciation dictionary used throughout these experiments is in British English \cite{beep}.

\section{Method overview}
We use the clustering approach of \cite{bear2014phoneme} to produce a series of P2V maps. We construct
\begin{enumerate}
\item a speaker-dependent map for each speaker;
\item a multi-speaker map using {\em all} speakers' phoneme confusions; 
\item a speaker-independent map for each speaker using confusions of all {\em other} speakers in the data. 
\end{enumerate}
Each P2V map is constructed using separate training and test data by using seven fold cross-validation \cite{efron1983leisurely}. In total each speaker utters 182 words (seven recitations of 26 words). Each one of seven recitations of the alphabet are selected as test folds in turn and are not included in the training folds.

We then use the HTK toolkit \cite{htk34} to build Hidden Markov Model (HMM) classifiers whose models are the viseme classes in each P2V map. We flat-start the HMMs with \texttt{HCompV}, re-estimate them 11 times over (\texttt{HERest}) with forced alignment between seventh and eighth re-estimates. Finally we recognise using \texttt{HVite} and output our results with \texttt{HResults}. The models are three state HMMs each having an associated Gaussian mixture of five components. Our recognition network constrains the output to be one of the 26 letters of the alphabet. Therefore, our measure of accuracy is $\dfrac{\#letters correct}{\#letters classified}$.

\section{Experimental setup}
We designate the P2V maps formed in these experiments as 
\begin{equation}
	M_n(p,q)\quad
        	\label{eq2}
\end{equation}
This means that the P2V map is derived from speaker $n$, but trained using visual speech data from speaker $p$ and tested using visual speech data from speaker $q$. For example, $M_1(2,3)$ would designate the result of testing a P2V map constructed from Speaker 1 using data from Speaker 2 to train the viseme models and testing on Speaker 3's data.  
\subsection{Baseline: Same Speaker Dependent maps (SSD)}
\label{sec:expSetup}
We establish a baseline of performance using the speaker-dependent results: $M_1(1,1), M_2(2,2), M_3(3,3)$ and $M_4(4,4)$. They are same speaker dependent (SSD) because the map, the models and the testing data are all derived from the same speaker.
Table~\ref{tab:sd} depicts how these maps are constructed. 
\begin{table}[!ht]
\centering
	\begin{tabular}{| l | l | l |}
	\hline
	\multicolumn{3}{| c |}{Same speaker-dependent (SD)} \\
	Mapping ($M_n$) & Training data ($p$) & Test speaker ($q$) \\
	\hline \hline
	Sp1 & Sp1 & Sp1 \\
	Sp2 & Sp2 & Sp2 \\
	Sp3 & Sp3 & Sp3 \\
	Sp4 & Sp4 & Sp4 \\
	\hline
	\end{tabular}
\caption{Same Speaker-Dependent (SSD) experiments used as a baseline for comparison}
\label{tab:sd}
\end{table}	
The resulting SSD P2V maps are listed in Table~\ref{tab:td_v}. The /garb/ viseme is made up of phonemes which did not appear in the output from the recogniser. Each viseme is listed with its associated mutually-confused phonemes e.g. for $M_1$, we see /v01/ is made up of phonemes \{/ah/, /iy/, /ow/, /uw/\}. These means in the phoneme recognition, all four phonemes \{/ah/, /iy/, /ow/, /uw/\} were confused with the other three in the viseme.

\subsection{Different Speaker Dependent maps \& Data (DSD\&D)}
In these tests, we use the HMM recognisers trained on each single speaker to recognise data from different speakers. This is done for all four speakers using the P2V maps of the other speakers, and the data from the other speakers. Hence for Speaker 1 we construct $M_2(2,1), M_3(3,1)$ and $M_4(4,1)$ and so on for the other speakers---this is depicted in Table~\ref{tab:sid}.  
\begin{table}[!ht]
\centering
	\begin{tabular}{| l | l | l |}
	\hline
	\multicolumn{3}{| c |}{Different Speaker Dependent maps \& Data (DSD\&D)} \\
	Mapping ($M_n$) & Training data ($p$) & Test speaker ($q$) \\
	\hline \hline
	Sp2,Sp3,Sp4 & Sp2,Sp3,Sp4 & Sp1 \\
	Sp1,Sp3,Sp4 & Sp2,Sp3,Sp4 & Sp2 \\
	Sp1,Sp2,Sp4 & Sp3,Sp2,Sp4 & Sp3 \\
	Sp1,Sp2,Sp3 & Sp4,Sp2,Sp3 & Sp4 \\
	\hline
	\end{tabular}
\caption{Different Speaker Dependent maps \& Data (DSD\&D) experiments}
\label{tab:sid}
\end{table}	

\begin{table*}[!ht]
\centering
\begin{tabular} {| l | l || l | l || l | l || l | l | }
	\hline
	\multicolumn{2}{| c ||}{Speaker 1 $M_1$} & \multicolumn{2}{| c ||}{Speaker 2 $M_2$} & \multicolumn{2}{| c ||}{Speaker 3 $M_3$} & \multicolumn{2}{| c |}{Speaker 4 $M_4$} \\
	Viseme & Phonemes 		& Viseme & Phonemes 		& Viseme & Phonemes			& Viseme & Phonemes \\
	\hline \hline
	/v01/	& /ah/ /iy/ /ow/ /uw/ 	& /v01/	& /ay/ /ey/ /iy/ /uw/		& /v01/	& /ey/ /iy/			& /v01/	& /ah/ /ay/ /ey/ /iy/	\\
	/v02/	& /ax/ /eh/ /ey/		& /v02/	& /ow/				& /v02/	& /ax/ /eh/			& /v02/	& /ax/ /eh/			\\
	/v03/	& /aa/ /ay/			& /v03/	& /ax/				& /v03/	& /ay/			& /v03/	& /aa/			\\
	/v04/	& /d/ /s/ /t/			& /v04/	& /eh/				& /v04/	& /ah/			& /v04/	& /ow/			\\
	/v05/	& /ch/ /l/  			& /v05/	& /ah/ 				& /v05/	& /aa/			& /v05/	& /uw/			\\
	/v06/	& /m/ /n/			& /v06/ 	& /aa/				& /v06/	& /ow/			& /v06/	& /m/ /n/			\\
	/v07/	& /jh/ /v/			& /v07/	& /jh/ /p/ /y/ 			& /v07/	& /uw/			& /v07/	& /k/ /l/			\\
	/v08/	& /b/ /y/ 			& /v08/	& /l/ /m/ /n/ 			& /v08/	& /d/ /p/ /t/			& /v08/	& /jh/ /t/			\\
	/v09/	& /k/				& /v09/	& /v/ /w/				& /v09/	& /l/ /m/			& /v09/	& /d/ /s/			\\
	/v10/	& /z/ 				& /v10/	& /d/ /b/				& /v10/	& /k/ /w/			& /v10/	& /w/				\\
	/v11/	& /w/				& /v11/	& /f/ /s/				& /v11/	& /f/ /n/			& /v11/	& /f/				\\
	/v12/	& /f/				& /v12/	& /t/ 					& /v12/	& /b/ /s/			& /v12/	& /v/				\\
		&				& /v13/	& /k/					& /v13/	& /v/				& /v13/	& /ch/			\\
		&				& /v14/	& /ch/				& /v14/	& /jh/				& /v14/	& /b/				\\
		&				&		&					& /v15/	& /ch/			& /v15/	& /y/				\\
		& 				&		&					& /v16/	& /y/				&		&				\\
		&				&		&					& /v17/	& /z/				& 		&				\\
	/sil/ 	& /sil/			& /sil/	& /sil/				& /sil/	& /sil/			& /sil/	& /sil/			\\
	/garb/& /ea/ /oh/ /ao/ /r/ /p/	& /garb/	& /ea/ /oh/ /ao/ /r/ /z/		& /garb/	& /ea/ /oh/ /ao/ /r/	& /garb/	& /ea/ /oh/ /ao/ /r/ /p/ /z/ \\

	\hline
\end{tabular}
\caption{Speaker-dependent phoneme-to-viseme mapping derived from phoneme recognition confusions for each speaker in AVL2}
\label{tab:td_v}
\end{table*}

\subsection{Different Speaker Dependent maps (DSD)}
 In our next experiment we train our models on speech from a single speaker but vary the speaker-dependent maps. This isolates the effects of the HMM recognition from the effect of different viseme classes. So for Speaker 1, we test the following Speaker-Independent Maps: $M_2(1,1), M_3(1,1)$ and $M_4(1,1)$. These are the same P2V maps in Table~\ref{tab:td_v} but trained and tested differently. This is depicted in Table \ref{tab:si}.
\begin{table}[!ht]
\centering
	\begin{tabular}{| l | l | l |}
	\hline
	\multicolumn{3}{| c |}{Different Speaker Dependent maps (DSD)} \\
	Mapping ($M_n$) & Training data ($p$) & Test speaker ($q$) \\
	\hline \hline
	Sp2,Sp3,Sp4 & Sp1 & Sp1 \\
	Sp1,Sp3,Sp4 & Sp2 & Sp2 \\
	Sp1,Sp2,Sp4 & Sp3 & Sp3 \\
	Sp1,Sp2,Sp3 & Sp4 & Sp4 \\
	\hline
	\end{tabular}
\caption{Different Speaker Dependent maps (DSD) experiments}
\label{tab:si}
\end{table}	

\subsection{Multi-speaker maps (MS)}
In the third set of experiments, we use the multi-speaker (MS) P2V map to form the viseme classes. This map is constructed using phoneme confusions produced by {\em all} our speakers and is shown in Table~\ref{tab:mt_v}. 
\begin{table}[!ht]
\centering
	\begin{tabular}{| l | l | l |}
	\hline
	\multicolumn{3}{| c |}{Multi-Speaker (MS)} \\
	Mapping ($M_n$) & Training data ($p$) & Test speaker ($q$) \\
	\hline \hline
	Sp1234 & Sp1 & Sp1 \\
	Sp1234 & Sp2 & Sp2 \\
	Sp1234 & Sp3 & Sp3 \\
	Sp1234 & Sp4 & Sp4 \\
	\hline
	\end{tabular}
\caption{Multi-Speaker (MS) experiments used as a baseline for comparison}
\label{tab:ms}
\end{table}	
We test this map as follows: $M_{[1234]}(1,1), M_{[1234]}(2,2), M_{[1234]}(3,3)$ and $M_{[1234]}(4,4)$: this is explained in Table~\ref{tab:ms}.   

\begin{table}[!ht]
\centering
\begin{tabular}{| l | l |}
	\hline
	\multicolumn{2}{| c |}{Multi-Speaker $M_{1234}$} \\
	Viseme & Phonemes \\
	\hline \hline
	/v01/ & /ah/ /ay/ /ey/ /iy/ /ow/ /uw/ \\
	/v02/ & /ax/ /eh/ \\
	/v03/ & /aa/ \\
	/v04/ & /d/ /s/ /t/ /v/ \\
	/v05/ & /f/ /l/ /n/ \\
	/v06/ & /b/ /w/ /y/ \\
	/v07/ & /jh/ \\
	/v08/ & /z/ \\
	/v09/ & /p/ \\
	/v10/ & /m/ \\
	/v11/ & /k/ \\
	/v12/ & /ch/ \\
	/sil/ & /sil/ \\
	/garb/ & /ea/ /oh/ /ao/ /r/ \\	\hline
\end{tabular}
\caption{Phoneme-to-viseme mapping derived from phoneme recognition confusions for all four speakers in AVL2}
\label{tab:mt_v}
\end{table}
	 
\subsection{Speaker-Independent maps (SI)}

Finally, we use our phoneme-clustering method to create a set of Speaker-Independent (SI) maps for each of the four speakers. These final P2V maps are shown in Table~\ref{tab:l1o_v}.
\begin{table}[!ht]
\centering
	\begin{tabular}{| l | l | l |}
	\hline
	\multicolumn{3}{| c |}{Speaker-Independent maps (SI)} \\
	Mapping ($M_n$) & Training data ($p$) & Test speaker ($q$) \\
	\hline \hline
	Sp234 & Sp1 & Sp1 \\
	Sp134 & Sp2 & Sp2 \\
	Sp124 & Sp3 & Sp3 \\
	Sp123 & Sp4 & Sp4 \\
	\hline
	\end{tabular}
\caption{Speaker-Independent (SI) maps experiments}
\label{tab:sim}
\end{table}
We test these maps as follows $M_{234}(1,1), M_{134}(2,2), M_{124}(3,3)$ and $M_{123}(4,4)$ as shown in Table \ref{tab:sim}.	

 \begin{table*}[!pht]
\centering
\begin{tabular}{| l | l || l | l || l | l || l | l | }
	\hline
	\multicolumn{2}{| c ||}{Speaker 1 $M_{234}$} & \multicolumn{2}{| c ||}{Speaker 2 $M_{134}$ } & \multicolumn{2}{| c ||}{Speaker 3 $M_{124}$ } & \multicolumn{2}{| c |}{Speaker 4 $M_{123}$ } \\
	Viseme & Phonemes 		& Viseme & Phonemes 		& Viseme & Phonemes			& Viseme & Phonemes \\
	\hline \hline
	/v01/ & 	/ah/ /ax/ /ay/		& /v01/	& 	/ah/ /ay/ /ey/ 	& /v01/ 	&	/ah/ /ay/ /ey/ 		& /v01/ 	& /ah/ /ay/ /ey/ \\
		&	 /ey/ /iy/			&		&	/iy/			&		&	/iy/ /ow/ /uw/ 		&		&  /iy/ /ow/ /uw/ \\
	v02 	& 	ow uw 			& v02 	& 	aa ow uw		& v02	& 	aa 				& v02	& aa \\	
	v03 	&	eh 				& v03 	& 	ax eh		& v03 	& 	ax eh			& v03	& ax eh \\
	v04 	& 	aa 				& v04 	& 	d s t 			& v04	& 	d s t v			& v04	& jh s t v\\	
	v05 	& 	d s t v 			& v05	& 	ch l 			& v05 	& 	l m n 			& v05	& f l n \\		
	v06 	& 	l m n 			& v06 	& 	b jh 			& v06 	& 	b w y 			& v06 	& b d p \\		
	v07 	& 	jh p y 			& v07	& 	v y			& v07	& 	jh 				& v07	& w y \\
	v08 	& 	k w 				& v08	& 	k w 			& v08	& 	z 				& v08	& z \\
	v09 	& 	f 				& v09 	& 	p			& v09 	&	p 				& v09 	& m \\
	v10 	& 	ch 				& v10 	& 	z 			& v10 	& 	k 				& v10 	& k \\
	v11 	& 	b 				& v11	& 	m			& v11	& 	f 				& v11 	& ch \\
		&					&		&				& v12	& 	ch				&  		&  \\			
	sil 	& 	sil 				& sil 		& 	sil			& sil 		& 	sil				& sil 		& sil \\
	garb & 	ea oh ao r z 		& garb 	& ea oh ao r f n		& garb	& 	ea oh ao r iy		& garb 	& ea oh ao r \\
	\hline
\end{tabular}
\caption{Phoneme-to-viseme mapping derived from phoneme recognition confusions of the three other speakers in AVL2}
\label{tab:l1o_v}
\end{table*}

  \subsection{Homophones}
  
\begin{table}[!ht]
\centering
\begin{tabular}{| l | r | } 
	\hline
	Map 	&	Unique words $T$ \\ 
	\hline \hline
	$M_1$	&	19		\\ 
	$M_2$	&	19		\\ 
	$M_3$	& 	24		\\ 
	$M_4$	& 	24		\\ 
	\hdashline
	$M_{1234}$	&	14	\\ 
	\hdashline
	$M_{234}$	&	17	\\ 
	$M_{134}$	&	18	\\ 
	$M_{124}$	&	20	\\ 
	$M_{123} $	&	15	\\ 
	\hline
\end{tabular}
\caption{Homophones created by each P2V mapping, allowing for variation in pronunciation} 
\label{tab:homophones}
\end{table}
Because the P2V maps are a many-to-one mapping, there is the
possibility of creating visual homophones. For example, the phonetic realisation of the word `B' is \textit{b iy} and of `D' is \textit{d iy}. Using map $M_2(2,2)$ they become B = \textit{v08 v01} and D = \textit{v08  v01} which are indistinguishable. The vocabulary of AVL2 is the 26 letters, A--Z. Permitting variations in pronunciation, we show the total unique words ($T$) for each map after each word (letter) has been translated from words, to phonemes, to visemes in Table~\ref{tab:homophones}. The higher the volume of homophones, the greater the chance of substitution errors.  

\section{Results}
\begin{figure*}[!ht]
        \centering
        \includegraphics[width=0.8\linewidth]{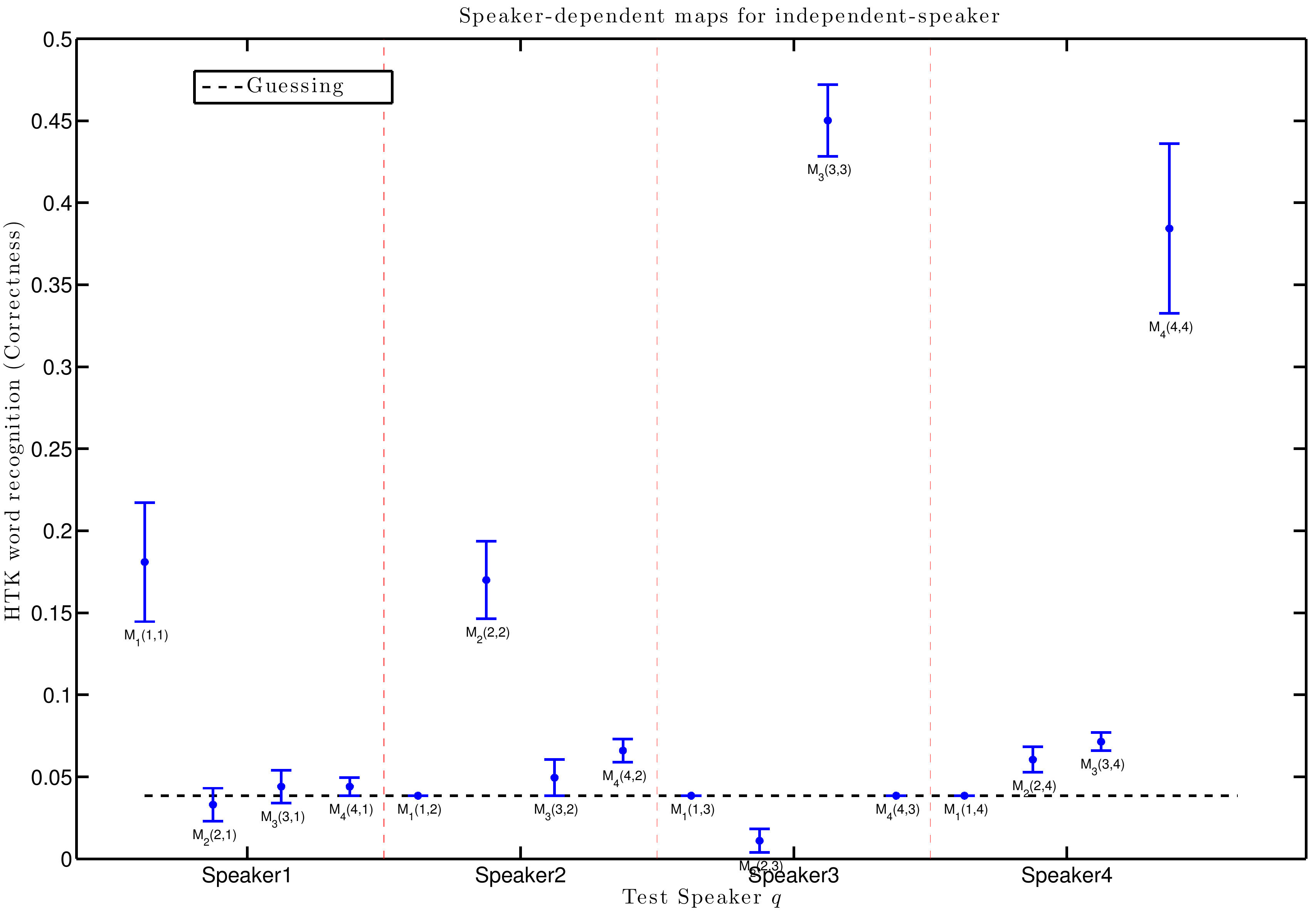}
        \caption{{\it Word recognition measured by correctness of the DSD\&D trained HMM classifiers used on all three other speakers in AVL2. Baseline is the SSD maps and error bars show $\pm$ one standard error.}}
        \label{fig:indep_Corr}
      \end{figure*}
      
Figure~\ref{fig:indep_Corr} shows the word recognition of speaker-dependent viseme classes, measured by correctness. In this figure, our baseline is $n=p=q$ for all $M$. We compare these to: $M_2(2,1), M_3(3,1), M_4(4,1)$ for Speaker 1, $M_1(1,2), M_3(3,2), M_4(4,2)$ for Speaker 2, $M_1(1,3), M_2(2,3), M_4(4,3)$ for Speaker 3 and $M_1(1,4), M_2(2,4), M_3(3,4)$ for Speaker 4. DSD HMM recognisers are significantly worse than SSD HMMs, as all results where $p$ is not the same speaker as $q$ are around the equivalent performance of  guessing. This correlates with similar tests of independent HMM's in \cite{cox2008challenge}. We can attribute this gap to two possible effects, either - the visual units are incorrect, or they are trained on the incorrect speaker. 

  \begin{figure*}[!ht]
        \centering
        \includegraphics[width=0.8\linewidth]{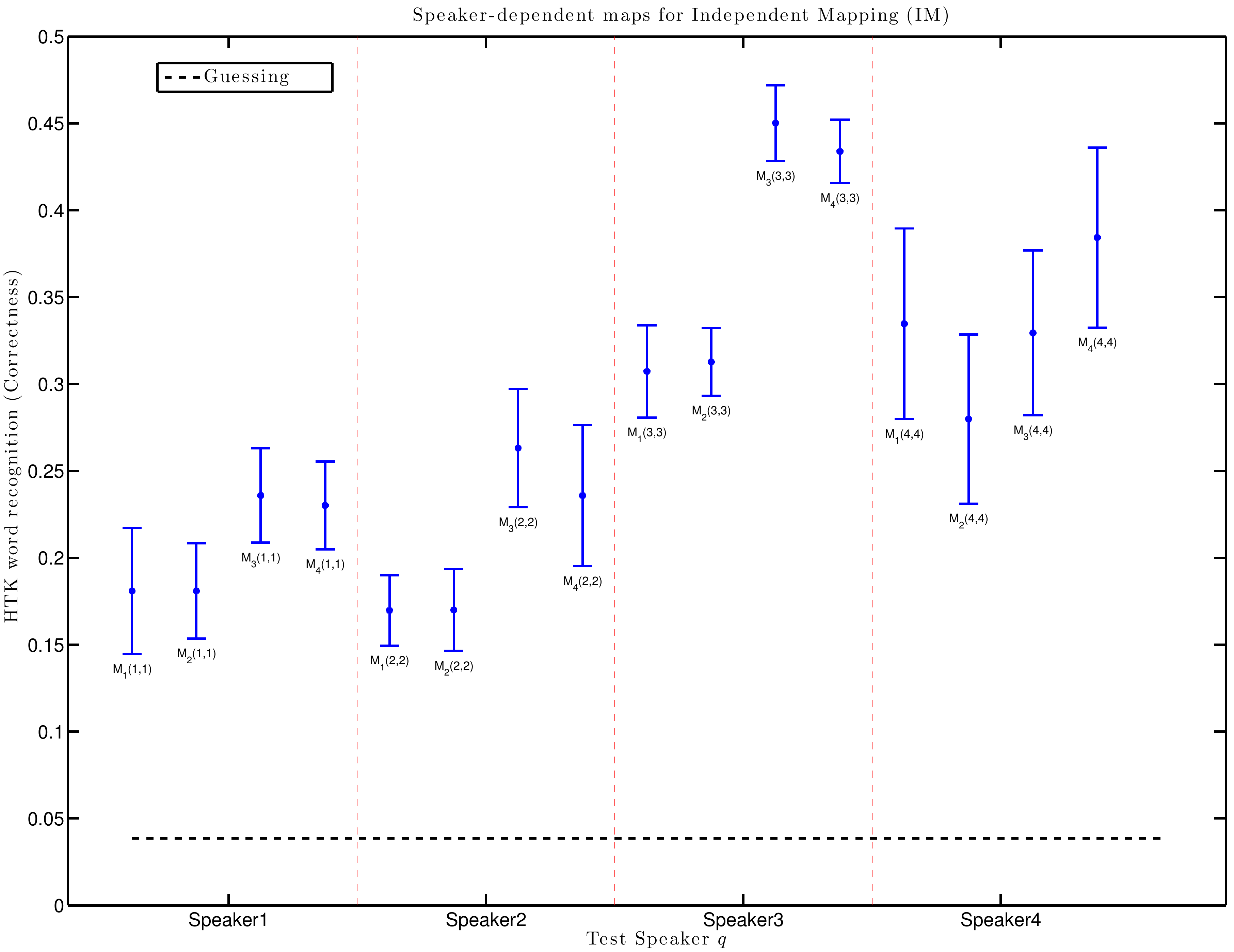}
        \caption{{\it Word recognition measured by correctness of the DSD classifiers constructed with single-speaker independent P2V maps for all four speakers in AVL2. Baseline is the SSD maps and error bars show $\pm$ one standard error.}}
        \label{fig:correctness}
      \end{figure*}
In Figure~\ref{fig:correctness} we have repeated the same benchmark as in Figure~\ref{fig:indep_Corr}($n=p=q$), but we have now allowed the HMM to be trained on the relevant speaker, so the other tests are: $M_2(1,1), M_3(1,1), M_4(1,1)$ for Speaker 1, $M_1(2,2,) M_3(2,2), M_4(2,2)$ for Speaker 2, $M_1(3,3), M_2(3,3), M_4(3,3)$ for Speaker 3 and finally $M_1(4,4), M_2(4,4), M_3(4,4)$ for Speaker 4. Now the word correctness has improved substantially which implies that the previous poor performance was not due to the choice of visemes but rather, the badly trained HMMs. 

We rank the performance of each viseme set on each speaker by weighting the effect of the DSD tests. We score each map as in Table~\ref{tab:weighting}. If a map increases on SSD performance within error bar range this scores $+1$ or outside error bar range scores $+2$. Likewise if a map decreases recognition on SSD performance, these values are negative. 
\begin{table}
\centering
	\begin{tabular}{|l|r|r|r|r|}
	\hline
		 & $M_1$ & $M_2$ & $M_3$ & $M_4$ \\
	\hline \hline
	Sp01 & $0$ 		& $+1$		& $+2$		& $+2$ \\
	Sp02 & $-1$		& $0$		& $+2$		& $+1$ \\
	Sp03 & $-2$		& $-2$		& $0$		& $-1$ \\
	Sp04 & $-1$		& $+1$		& $-1$		& $0$ \\
 	Total	 & $-4$		& $0$		& $3$		& $2$ \\
	\hline
	\end{tabular}
	\caption{Weighted scores from comparing the use of speaker-dependent maps for \emph{other} speaker-dependent lip-reading}
	\label{tab:weighting}
\end{table}

So we see that $M-3$ is the best of the four SSD maps, followed by $M_4$, $M_2$ and finally $M_1$ is the most susceptible to speaker identity. We note that this order matches a decreasing order of quantity of visemes in the speaker-dependent viseme sets i.e. the more similar to phoneme classes visemes are, then the better the recognition performance. This ties in with Table~\ref{tab:homophones}, where the better P2V maps have less homorphous words. 


In Table~\ref{tab:td_v}, phoneme pairs \{/ax/, /eh/\}, \{/m/, /n/\} and \{/ey/, /iy/\} are present for three speakers and \{/ah/, /iy/\} and \{/l/, /m/\} are pairs for two speakers. Of the single-phoneme visemes, /ch/ is present three times, /f/, /k/, /w/ \& /z/ twice. 

The important lesson from Figure~\ref{fig:correctness}, is that the selection of incorrect units, whilst detrimental, is not as devastating as training recognition classes on alternative speakers. 

       \begin{figure*}[!t]
        \centering
        \includegraphics[width=0.8\linewidth]{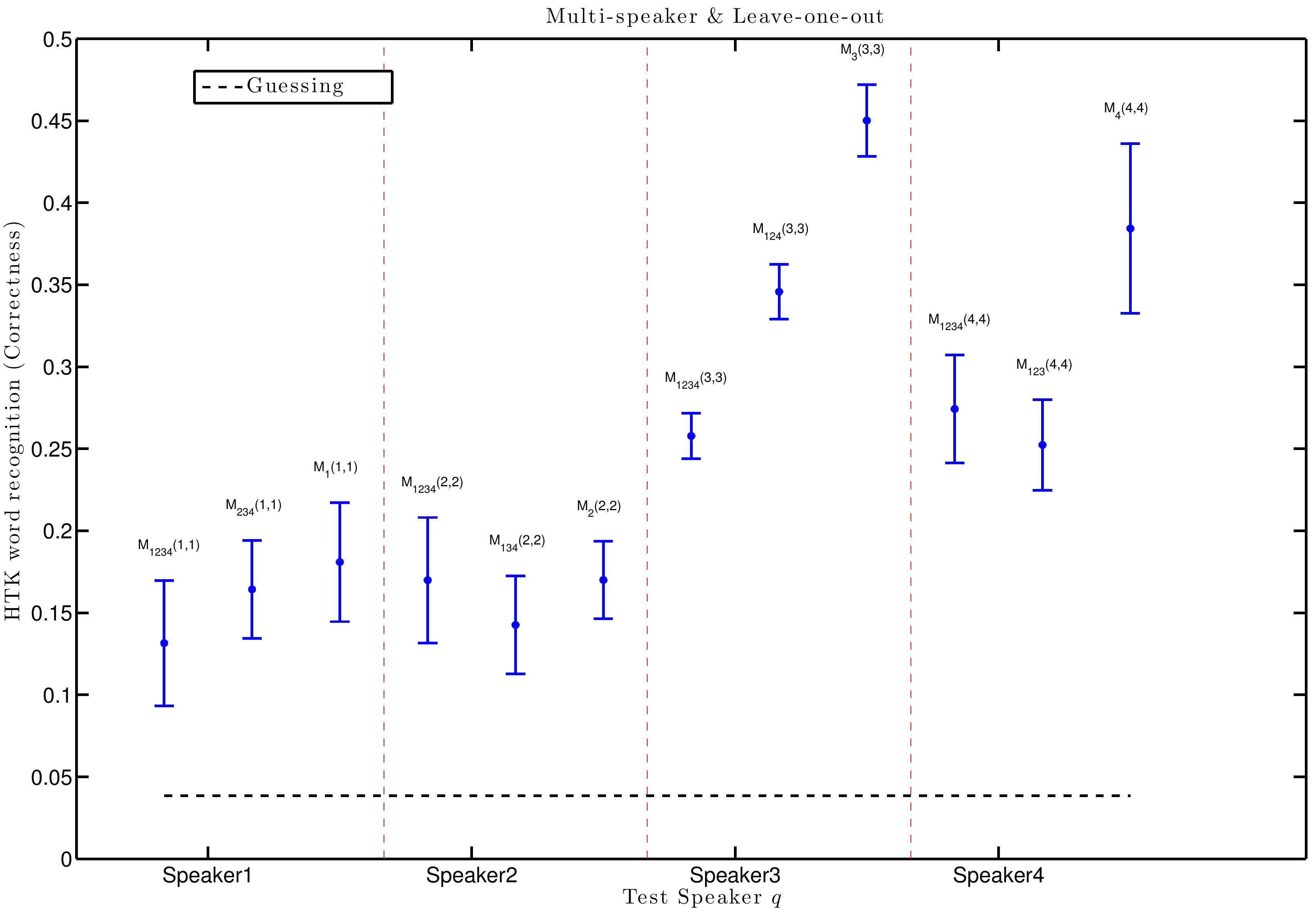}
        \caption{{\it Word recognition measured by correctness of the classifiers using MS and SI phoneme-to-viseme maps. Baseline is the SSD maps and error bars show $\pm$ one standard error.}}
        \label{fig:accuracy}
      \end{figure*}
      
  Figure~\ref{fig:accuracy} shows the correctness of both the MS viseme class set and the SI sets. For the multi-speaker classifiers, these are all built on the same map $M_{1234}$, and tested on the same speaker so, $p=q$. Therefore our tests are: $M_{1234}(1,1), M_{1234}(2,2), M_{1234}(3,3), M_{1234}(4,4)$. To test our SI maps, we plot  $M_{234}(1,1), M_{134}(2,2), M_{124}(3,3)$ and $M_{123}(4,4)$. Again we repeat the same baseline where $n=p=q$ for reference. 
  
There is no significant difference on Speaker 2, and while Speaker 3 word recognition is reduced, it is not eradicated. It is interesting that for Speaker 3, for whom their speaker-dependent recognition was the best of all speakers, the SIM map ($M_{124}$) out performs the multi-speaker viseme classes ($M_{1234}$) significantly. This maybe due to Speaker 3 having a unique visual talking style which reduces similarities with Speakers 1, 2 \& 4. 
  
 If we compare all the P2V maps in Tables~\ref{tab:mt_v} \&~\ref{tab:l1o_v}, there are similarities. Mostly because we know there is only one speaker at a time removed from within SIM P2V maps. However, if we compare these to the speaker-dependent maps in Table~\ref{tab:td_v}, we see a different picture. Speaker 4 is significantly affected by the introduction of /ow/ and /uw/ into viseme /v01/. Where Speaker 1 has these in $M_1(1,1)$, we see that his SD word recognition of $15.9\%$ is less than half of Speaker 4's $38.4\% $ (Figure~\ref{fig:correctness}). 
   
 \section{Conclusions}

Our principal conclusion can be seen by comparing Figures~\ref{fig:correctness} \&~\ref{fig:accuracy} with Figure~\ref{fig:indep_Corr}. Figure~\ref{fig:indep_Corr} shows a very substantial reduction in performance when the system is truing on a speaker who is not the test speaker. The question arises as to whether this degradation is due to the wrong choice of map or the wrong training data for the recognisers. We conclude that is  it not the choice of map that causes degradation since we can retrain the HMMs and regain much of the performance. We regain performance irrespective of whether the map is chosen for a different speaker, multi-speaker or independently of the speaker. 

This is an important conclusion since it tells us that the repertoire of lip appearances does not vary significantly across speakers. This is comforting since the prospect of recognition using a symbol alphabet which varies by speaker is daunting. This is further reinforced by Tables~\ref{tab:td_v},~\ref{tab:mt_v} \&~\ref{tab:l1o_v}. There are differences between speakers, but not significant ones. 

An analogy with acoustic speech would be the question of whether an accented Norfolk speaker requires a different set of phonemes to a standard British talker. No: they can be represented by the same set of phonemes; they just use these phonemes in a different way.




  \eightpt 
  \bibliographystyle{IEEEtran}

  \bibliography{mybib}

\end{document}